\documentclass[11pt,a4paper]{article}
\usepackage[hyperref]{emnlp2020}
\usepackage{times}
\usepackage{latexsym}

\usepackage{graphicx}
\usepackage{algorithm}
\usepackage{algpseudocode}
\usepackage{multirow}
\usepackage{multicol}
\usepackage{makecell}
\usepackage{caption}
\usepackage{subcaption}
\usepackage{diagbox}
\usepackage{amssymb}
\usepackage{booktabs}

\usepackage{amsmath,amsfonts,bm}

\def\eqref#1{equation~\ref{#1}}

\def\1{\bm{1}}

\DeclareMathAlphabet{\mathsfit}{\encodingdefault}{\sfdefault}{m}{sl}
\SetMathAlphabet{\mathsfit}{bold}{\encodingdefault}{\sfdefault}{bx}{n}

\usepackage{microtype}

\aclfinalcopy 
\def\aclpaperid{2124} 

\title{Improving Word Embedding Factorization for Compression Using Distilled Nonlinear Neural Decomposition}

\author{
 Vasileios Lioutas\textsuperscript{1}\thanks{$\;\;$Work done during an internship at Huawei Noah’s Ark Lab.} , Ahmad Rashid\textsuperscript{2}, Krtin Kumar\textsuperscript{2}, \\ \textbf{Md. Akmal Haidar\textsuperscript{2}, Mehdi Rezagholizadeh\textsuperscript{2}} \\
 \textsuperscript{1}University of British Columbia, \textsuperscript{2}Huawei Noah’s Ark Lab\\
 \texttt{\normalsize contact@vlioutas.com,} \texttt{\normalsize ahmad.rashid@huawei.com,} \texttt{\normalsize krtin.kumar@huawei.com,} \\
 \texttt{\normalsize md.akmal.haidar@huawei.com,} \texttt{\normalsize mehdi.rezagholizadeh@huawei.com}
}

\date{}

\begin{document}
\maketitle
\begin{abstract}
Word-embeddings are vital components of Natural Language Processing (NLP) models and have been extensively explored. However, they consume a lot of memory which poses a challenge for edge deployment. Embedding matrices, typically, contain most of the parameters for language models and about a third for machine translation systems. In this paper, we propose Distilled Embedding, an (input/output) embedding compression method based on low-rank matrix decomposition and knowledge distillation. First, we initialize the weights of our decomposed matrices by learning to reconstruct the full pre-trained word-embedding and then fine-tune end-to-end, employing knowledge distillation on the factorized embedding. We conduct extensive experiments with various compression rates on machine translation and language modeling, using different data-sets with a shared word-embedding matrix for both embedding and vocabulary projection matrices. We show that the proposed technique is simple to replicate, with one fixed parameter controlling compression size, has higher BLEU score on translation and lower perplexity on language modeling compared to complex, difficult to tune state-of-the-art methods.
\end{abstract}

\section{Introduction}
Deep Learning models are the state-of-the-art in NLP, Computer Vision, Speech Recognition and many other fields in Computer Science and Engineering. The remarkable deep learning revolution has been built on top of massive amounts of data (both labeled and unlabeled), and faster computation. In NLP, large pre-trained language models like BERT \cite{devlin2018bert} are state-of-the-art on a large number of downstream NLP problems. The largest publicly available language models are trained with hundred of billions of parameters \cite{brown2020language}. In machine translation the state-of-the-art models have parameters in the order of billions. Data privacy and server cost are some major issues, driving research towards deploying these models on edge-devices. However, running these models on edge-devices, faces memory and latency issues due to limitations of the hardware. Thus, there has been considerable interest towards research in reducing the memory footprint and faster inference speed for these models \cite{sainath2013low,Acharya2019,shi2018structured,jegou2010product,chen2018groupreduce,winata2018low}.

The architecture of deep-learning-based language generation models can be broken down into three components. The first component, represents the embedding, which maps words in the vocabulary to continuous dense vector representations of the words. In language modeling we typically have one dictionary but machine translation has at least two dictionaries corresponding to a translation pair. We model these as a single dictionary with a common embedding matrix. The second component, consists of a function $f$, typically a deep neural-network \cite{schmidhuber2015deep,krizhevsky2012imagenet,mikolov2010recurrent} which maps the embedding representation for different NLP problems (machine-translation, summarization, question-answering and others), to the output-space of function $f$. The third component, is the output layer which maps the output of function $f$ to the vocabulary-space, followed by a softmax function. Since, the first and third components depend upon a large vocabulary-size, they require large number of parameters which results in higher latency and larger memory requirements. For instance, the Transformer Base model \cite{NIPS2017_7181} uses 37\% of the parameters in the first and third components using a vocabulary size of 50k, and with parameter-tying between the components. The percentage of parameters increases to 54\%, when parameters are not shared between the first and third components. Thus, an obvious step towards model compression is to reduce the parameters used by the embedding matrices.

Recently, there has been considerable work on compressing word-embedding matrices \cite{sainath2013low,Acharya2019,shi2018structured,jegou2010product,chen2018groupreduce,winata2018low}. These techniques have proven to perform at-par with the uncompressed models, but still suffer from a number of issues.

\textbf{First}, state-of-the-art embedding compression methods such as GroupReduce, Structured Emebedding and Tensor Train Decomposition \cite{shi2018structured,chen2018groupreduce,khrulkov2019tensorized,shu2018compressing}, require multiple hyper-parameters to be fine-tuned to optimize performance on each dataset. These hyper-parameters influence the number of parameters in the model, and thus the compression rate. This leads to an additional layer of complexity for optimizing the model for different NLP problems. Additionally, \citet{chen2018groupreduce} requires an additional optimization step for grouping words, and lacks end-to-end training through back-propagation. \citet{shi2018structured} also requires an additional step for performing k-means clustering for generating the quantization matrix. Thus, most of the current state-of-the-art systems are much more complicated to fine-tune for different NLP problems and data-sets.

\textbf{Second}, all the state-of-the-art embedding compression models compress the input and output embedding separately. In practice, state-of-the-art NLP models \cite{NIPS2017_7181,time-awareconv} have shown better performance with parameter sharing between the two \cite{press2016using}. Thus, there is a need for an exhaustive analysis of various embedding compression techniques, with parameter sharing.

\textbf{Lastly}, embedding compression models not based on linear SVD \cite{khrulkov2019tensorized,shi2018structured} require the reconstruction of the entire embedding matrix or additional computations, when used at the output-layer. Thus during runtime, the model either uses the same amount of memory as the uncompressed model or pays a higher computation cost. This makes linear SVD based techniques more desirable for running models on edge-devices.  

In this paper, we introduce Distilled Embedding, a matrix factorization method, based on Singular Value Decomposition (SVD) with two key changes a) a neural network decomposition instead of an eigenvalue decomposition and b) a distillation loss on the word embedding while fine-tuning. Our method, first compresses the vocabulary-space to the desired size, then applies a non-linear activation function, before recovering the original embedding-dimension. Additionally, we also introduce an embedding distillation method, which is similar to Knowledge Distillation \cite{Hinton2015} but we apply it to distill knowledge from a pre-trained embedding matrix and use an $L2$ loss instead of cross-entropy loss. To summarize, our contributions are as follows:   
\begin{itemize}
    \item We demonstrate that SVD, when fine-tuned till convergence, is comparable to recently proposed, difficult to tune methods.
    \item We demonstrate that at the same compression rate Distilled Embedding outperforms existing state-of-the-art methods on machine translation and SVD on language modeling.
    \item Our proposed method is much simpler than the current state-of-the-methods, with only a single parameter controlling the compression rate.
    \item Unlike the current state-of-the-art systems, we compress the embedding matrix with parameter sharing between input and output embeddings. We perform an exhaustive comparison of different models in this setting.
    \item Our method is faster at inference speed than competing matrix factorization methods and only slightly slower than SVD. 
\end{itemize}

\section{Related Work}
\begin{figure*}[t]
\centering
\begin{subfigure}[b]{0.51\textwidth}
\includegraphics[width=0.92\textwidth]{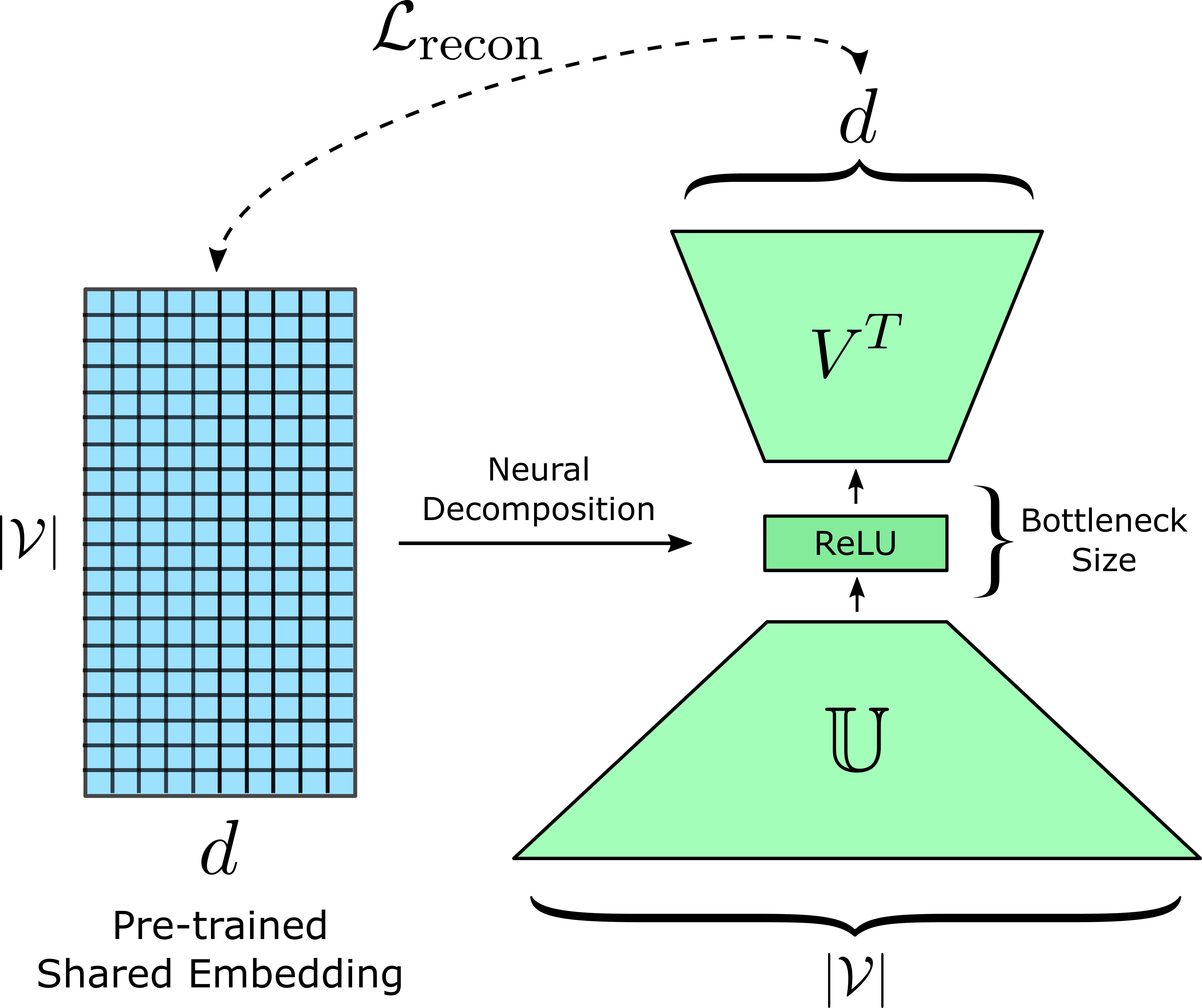}
\caption{Funneling Decomposition}
\label{fig:funneling}
\end{subfigure}%
\begin{subfigure}[b]{0.49\textwidth}
\centering
\includegraphics[width=1\textwidth]{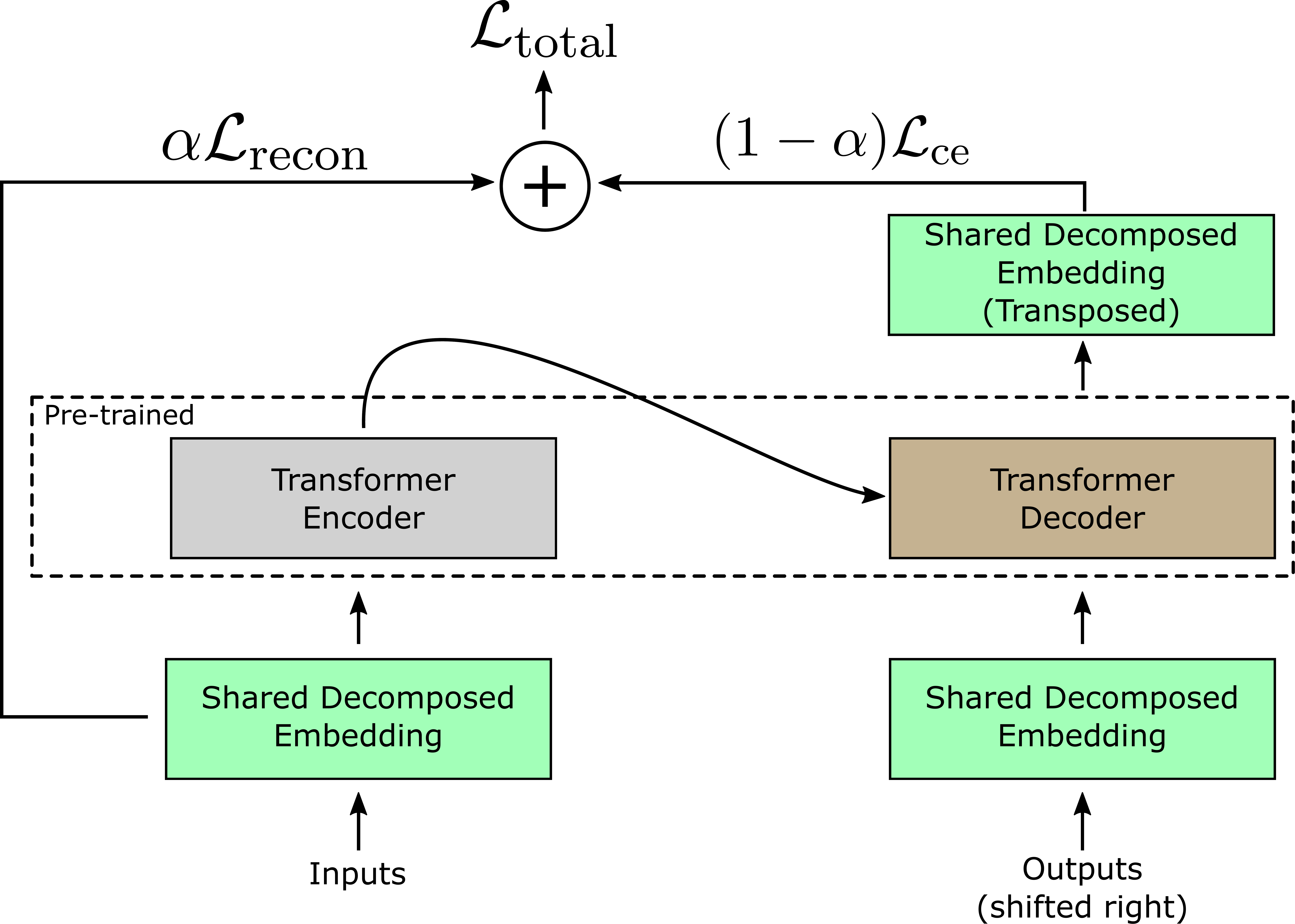}
\caption{Finetuning all the weights with knowledge distillation on the embedding}
\label{fig:seq2seq_arch}
\end{subfigure}
\caption{Distilled Embedding method to compress the shared embedding matrix of a transformer based sequence to sequence model.}
\label{fig:archs}
\end{figure*}

We can model the problem of compressing the embedding matrix as a matrix factorization problem. There is a considerable amount of work done in this field and some of the popular methods include Singular Value Decomposition (SVD) \cite{srebro2003weighted,mnih2008probabilistic}, product quantization \cite{jegou2010product} and tensor decomposition \cite{de2000multilinear}. A number of prior works in embedding compression are influenced by these fields and have been applied to various NLP problems. In this Section, we will discuss some of the significant works across different NLP problems.

\paragraph{Low-rank Factorization}
Low-rank approximation of weight matrices, using SVD, is a natural way to compress deep learning based NLP models. \citet{sainath2013low} apply this to a convolutional neural network for language modeling and acoustic modeling. \citet{winata2018low} use SVD on all the weight matrices of an LSTM and demonstrate competitive results on question-answering, language modeling and text-entailment. \citet{Acharya2019} use low-rank matrix factorization for word-embedding layer during training to compress a classification model. However, they do not study the effects of applying a non-linear function before reconstructing the original dimension.

\paragraph{GroupReduce}\citet{chen2018groupreduce} apply weighted low-rank approximation to the embedding matrix of an LSTM. They first create a many-to-one mapping of all the words in the vocabulary into $g$ groups based upon word frequency. For each group $g$ they apply weighted SVD to obtain a lower rank estimation, the rank is determined by setting a minimum rank and linearly increasing it based upon average frequency. Finally, they update the groups by minimizing the reconstruction error from the weighted SVD approximation. They demonstrate strong results on language modeling and machine translation compared to simple SVD. In their models they use different embedding matrices for input and softmax layers and apply different compression ratios to each.

\paragraph{Product Quantization} \citet{jegou2010product} introduced product quantization for compressing high dimensional vectors, by uniformly partitioning them into subvectors and quantizing each subvector using K-means clustering technique. Basically, product quantization assumes that the subvectors share some underlying properties which can be used to group similar ones together and unify their representation. That being said, this approach breaks the original matrix into a set of codebooks coming from the center of the clusters in different partitions together with a separate index matrix which refers to the index of the clusters for each subvector. \citet{shi2018structured} applied product quantization to a language model and were able to show better perplexity scores. \citet{shu2018compressing} extended this technique by first representing the product quantization as a matrix factorization problem, and then learning the quantization matrix in an end-to-end trainable neural network. \citet{li2018slim} implement product quantization through randomly sharing parameters in the embedding matrix, and show good results on perplexity for an LSTM based language model.

\paragraph{Tensor Decomposition} \citet{de2000multilinear} introduced multilinear SVD, which is a generalization of SVD for higher order tensors. \citet{oseledets2011tensor} introduced an efficient algorithm Tensor Train (TT) for multilinear SVD Tensor. \citet{novikov2015tensorizing} applied the Tensor Train decomposition on fully connected layers of deep neural networks. \citet{khrulkov2019tensorized} applied Tensor Train algorithm to the input embedding layer on different NLP problems like language modeling, machine translation and sentiment analysis. They demonstrate high compression rate with little loss of performance. However, they compress only the input embedding and not the softmax layer for language modeling and machine translation.  

\paragraph{Knowledge Distillation}
Knowledge distillation \cite{bucilua2006model,Hinton2015}. has been studied in model compression where knowledge of a large cumbersome model is transferred to a small model for easy deployment. In this paper, we propose an embedding factorization of word-embedding matrix using knowledge distillation to mimic the pre-trained word-embedding representation.

\section{Methodology: Distilled Embedding}

\subsection{Funneling Decomposition and Embedding Distillation}

We present an overview of our proposed method in Figure \ref{fig:archs}. Given an embedding matrix $E \in \mathbb{R}^{|\mathcal{V}| \times d}$, we can decompose it into three matrices (Equation \ref{eqn:svd}), using the SVD algorithm
\begin{equation}
    E = U_{|\mathcal{V}| \times |\mathcal{V}|} \Sigma_{|\mathcal{V}| \times d} V^T_{d \times d} 
    \label{eqn:svd}
\end{equation}
where $|\mathcal{V}|$ is the vocabulary size and $d$ is the embedding dimension. $\Sigma$ is a diagonal matrix containing the singular values, and matrices $U$ and $V$ represent the left and right singular vectors of the embedding matrix respectively. We can obtain the reduced form of the embedding matrix, $\tilde{E}$, by only keeping $r$ ($<d$) largest singular values out of $d$.        
\begin{equation}
    \tilde{E} = U_{|\mathcal{V}| \times r} \Sigma_{r \times r} V^T_{r \times d} = \mathbb{U}_{|\mathcal{V}| \times r}V^T_{r \times d}
\end{equation}
where the matrix $\mathbb{U} = U \Sigma$. The reduced form of the embedding matrix will need $r \times (|\mathcal{V}|+d)$ parameters compared to $|\mathcal{V}| \times d$. 

Our proposed approach in this work, is to apply a non-linear transformation on the matrix $\mathbb{U}$, before reconstructing the original embedding dimension using $V$ (see Figure 1a), as shown in Equation \ref{eq:our_method},
\begin{equation}
    \tilde{E} = f(\mathbb{U}_{|\mathcal{V}| \times r}) V^T_{r \times d}
    \label{eq:our_method}
\end{equation}

We use the ReLU as our non-linear function $f(.)$ throughout this paper. We postulate that this neural decomposition helps in end-to-end training during the fine-tuning stage, although, we can only demonstrate empirical evidence for that. We train a sequence to sequence model \cite{DBLP:journals/corr/SutskeverVL14,NIPS2017_7181} with tied input and output embedding (i.e.  the output embedding is the transpose of the input embedding matrix $\tilde{E}_{\text{out}}=\tilde{E}^T = V_{d \times r}[f(\mathbb{U}_{|\mathcal{V}| \times r})]^T. $
We train our model end-to-end by replacing the embedding function with Equation \ref{eq:our_method}. The matrix $\mathbb{U}$ and $V$ are trainable parameters, and for the output layer we use $\tilde{E}^T$, with the parameter sharing. We train on two losses. The standard cross entropy loss defined as:
\begin{equation}
    \mathcal{L}_{\textnormal{ce}} = -\sum_{i=1}^{M} \textnormal{y}_{i}\textnormal{log}(p_{i})
    \label{eq:ce}
\end{equation}
where $M$ is the sequence length, $\textnormal{y}_i$ is the one-hot representation for the  $i^{\text{th}}$ label and $p_i$ is the softmax probability of the  $i^{\text{th}}$ term generated by the decoder.

In addition to the cross-entropy loss, we introduce a novel embedding reconstruction loss (Equation \ref{eq:recon}), which we refer to as embedding distillation as we distill information from the pre-trained embedding into our model,
\begin{equation}
\begin{aligned}
    \mathcal{L}_{\textnormal{recon}} &= \frac{1}{\lvert \mathcal{V} \rvert}\sum_{i=1}^{\lvert \mathcal{V} \rvert}\| e_{i}-\tilde{e}_{i}\|_2 \\ &= \frac{1}{| \mathcal{V} |}\sum_{i=1}^{| \mathcal{V} |}\|e_{i}- f(u_i)  V^T_{r \times d}\|_2
    \label{eq:recon}
\end{aligned}
\end{equation}

where $e_i$ and $\tilde{e}_i$ are the embedding vectors corresponding to the $i^{\text{th}}$ word in the original embedding matrix $E$ and the reconstructed embedding matrix $\tilde{E}$ respectively and $u_i$ refers to the $i^{\text{th}}$ row of the matrix $\mathbb{U}$. This helps in better generalization since during fine-tuning the words seen in the training corpus are given higher weight at the expense of low-frequency word. This loss helps maintain a balance between the two. 

We use Equation \ref{eq:total} as our final loss function

\begin{equation}
    \mathcal{L}_{\textnormal{total}} = \alpha\mathcal{L}_{\textnormal{recon}} + (1-\alpha)\mathcal{L}_{\textnormal{ce}}
    \label{eq:total}
\end{equation}
where $\alpha \in [0,1]$ is a hyper-parameter, which controls the trade-off between reconstruction and cross-entropy loss. $\mathcal{L}_{\textnormal{recon}}$ acts as the knowledge distillation loss by which we try to distill information from the original pre-trained embedding layer as a teacher to the funneling decomposed embedding layer as a student. The training process of our Distilled Embedding method is summarized in \textbf{Algorithm \ref{alg}}.

\begin{algorithm}[t]
\caption{Distilled Embedding}
\label{alg}
\begin{algorithmic}
\State \textbf{Step 1) Pre-training the Embedding Matrix} Pre-train the sequence to sequence model with the full embedding matrix for better initialization.
\State \textbf{Step 2) Initializing the Weights of Funneling Decomposition Layer} We extract the trained embedding matrix $E$ from Step 1 and train our decomposed matrices $\mathbb{U}$ and $V$ on reconstruction loss defined in Equation \ref{eq:recon}, as shown in Figure~\ref{fig:funneling}. 
\State \textbf{Step 3) Embedding Distillation} The pre-trained funneling decomposition layer is plugged into the model (replacing the original embedding matrix $E$) and the entire model is trained based on Equation \ref{eq:total}. 
\end{algorithmic}
\end{algorithm}

\section{Experimental Setup}

\subsection{Datasets and Evaluation}
\label{sec:setup}
We test our proposed method on machine translation and language modeling which are fundamental problems in NLP and challenging for embedding compression since we typically have an input and output embedding.

On machine translation, we present results on three language pairs: WMT English to French (En-Fr), WMT English to German (En-De) and IWSLT Portuguese to English (Pt-En). We decided that these pairs are good representatives of high-resource, medium-resource and low-resource language pairs. 

WMT En-Fr is based on WMT’14 training data which contains 36M sentence pairs. We used SentencePiece \cite{DBLP:journals/corr/abs-1808-06226} to extract a shared vocabulary of 32k subwords. We validate on newstest2013 and test on newstest2014. For WMT English to German (En-De), we use the same setup as \citet{NIPS2017_7181}. The dataset is based on WMT’16 training data and contains about 4.5M pairs. We use a shared vocabulary of 37k subwords extracted using SentencePiece.

For the IWSLT Portuguese to English (Pt-En) dataset, we replicate the setup of \citet{MNMT_KD} for training individual models. Specifically, the dataset contains about 167k training pairs. We used a shared vocabulary of 32k subwords extracted with SentencePiece.

For all language pairs, we measure case-sensitive BLEU score \cite{papineni2002bleu} using SacreBLEU\footnote{\url{https://github.com/mjpost/sacreBLEU}} \cite{post-2018-call}. In addition, we save a checkpoint every hour for the WMT En-Fr and WMT En-De language pairs and every 5 minutes for the IWSLT Pt-En due to the smaller size of the dataset. We use the last checkpoint which resulted in the highest validation BLEU and average the last five checkpoints based on this. We use beam search with a beam width of 4 for all language pairs.

For language modeling, we decided to use the WikiText-103 dataset \cite{DBLP:journals/corr/MerityXBS16} which contains 103M training tokens from 28K articles, with an average length of 3.6K tokens per article. We replicate the setup of \citet{dai2019transformerxl} for training the base and the compressed models.

\begin{table*}[t]
\centering
\begin{tabular}{l|cc|cc|cc}
\toprule
\multirow{2}{*}{Model} & \multicolumn{2}{c|}{WMT En-Fr} & \multicolumn{2}{c|}{WMT En-De} & \multicolumn{2}{c}{IWSLT Pt-En} \\
\cline{2-7}
& \makecell{Emb. \\ CR} & \makecell{BLEU} & \makecell{Emb. \\ CR} & \makecell{BLEU} & \makecell{Emb. \\ CR} & \makecell{BLEU} \\
\midrule
Transformer Base & 1.0x & 38.12 & 1.0x & 27.08 & 1.0x & 41.43 \\
\midrule
Smaller Transformer Network (416) & 1.23x & 37.26 & 1.28x & 26.72 & 
1.88x & 40.71 \\
End-to-End NN compression with non-linearity & 7.87x & 37.23 & 7.89x & 26.14 & 3.96x & 42.27 \\
SVD with rank 64 & 7.87x & 37.44 & 7.89x & 26.32 & 3.96x & 42.37 \\
GroupReduce \cite{chen2018groupreduce} & 7.79x & 37.63 & 7.88x & 26.75 & 3.96x & 42.13 \\
Structured Embedding \cite{shi2018structured} & 7.90x & \textbf{37.78} & 7.89x & 26.34 & 3.97x & 41.27 \\
Tensor Train \cite{khrulkov2019tensorized} & 7.72x & 37.27 & 7.75x & 26.19 & 3.96x & 42.34 \\
\midrule
Distilled Embedding (Ours) & 7.87x & \textbf{37.78} & 7.89x & \textbf{26.97} & 3.96x & \textbf{42.62} \\
\bottomrule
\end{tabular}
\caption{Machine translation BLEU score for the three language pairs. CR refers to the compression rate.}
\label{tab:mtfr}
\end{table*}

\subsection{Experiment Details}

\paragraph{Hyper-Parameters} For WMT En-Fr and WMT En-De, we use the same configuration as Transformer Base which was proposed by \citet{NIPS2017_7181}. Specifically, the model hidden size $d_{\text{model}}$ is set to 512, the feed-forward hidden size $d_{\text{ff}}$ is set to 2048 and the number of layers for the encoder and the decoder was set to 6. For the IWSLT Pt-En, we use Transformer Small configuration. Specifically, the model hidden-size $d_{\text{model}}$ is set to 256, the feed-forward hidden size $d_{\text{ff}}$ is set to 1024 and the number of layers for the encoder and the decoder was set to 2. For Transformer Small, the dropout configuration was set the same as Transformer Base. All models are optimized using Adam \cite{DBLP:journals/corr/KingmaB14} and the same learning rate schedule as proposed by \citet{NIPS2017_7181}. We use label smoothing with 0.1 weight for the uniform prior distribution over the vocabulary \cite{DBLP:journals/corr/SzegedyVISW15,DBLP:journals/corr/PereyraTCKH17}. Additionally, we set the value $\alpha$ of Equation \ref{eq:total} to 0.01.

For the WikiText-103 we use the same configuration as Transformer-XL Standard which was proposed by \citet{dai2019transformerxl}. Specifically, the model hidden size $d_{\text{model}}$ is set to 410, the feed-forward hidden size $d_{\text{ff}}$ is set to 2100 and the number of layers for was set to 16.

\paragraph{Hardware Details} We train the WMT models on 8 NVIDIA V100 GPUs and the IWSLT models on a single NVIDIA V100 GPU. Each training batch contained a set of sentence pairs containing approximately 6000 source tokens and 6000 target tokens for each GPU worker. All experiments were run using the TensorFlow framework\footnote{\url{https://www.tensorflow.org/}}.

\section{Results}
\subsection{Machine Translation}
We present BLEU score for our method and compare it with SVD, GroupReduce \cite{chen2018groupreduce}, Structured Emedding \cite{shi2018structured}, Tensor Train \cite{khrulkov2019tensorized} and a smaller transformer network with the same number of parameters. We learn a decomposition for all the methods except Tensor Train since it was pointed out in \citet{khrulkov2019tensorized} that there is no difference in performance between random initialization and tensor train learnt initialization. Once initialized we plug the decomposed embedding and fine-tune till convergence. None of the weights are frozen during fine-tuning.

Table~\ref{tab:mtfr} presents the results on translation. We see that on the English-French language pair our method along with Structured Embedding performs the best. Group Reduce is next, and SVD performs better than Tensor Train, showing that SVD is a strong baseline, when fine-tuned till convergence. We also compare against end-to-end compression using a 2 layer neural network (NN) with the same parameterization as distilled embedding which has not been initialized offline. The results show that initializing the neural decomposition with the embedding weights is important. 

On English-German translation, our method outperforms all other methods. The smaller transformer network does well and is only surpassed by GroupReduce amongst the competing methods. SVD again performs better than Tensor Train.

\begin{table}[t]
\centering
\begin{tabular}{lccc}
\toprule
Model & \makecell{Emb. \\ CR} & \makecell{Val.\\PPL} & \makecell{Test\\PPL} \\
\midrule
\makecell[l]{Transformer-XL std\\\cite{dai2019transformerxl}} & 1.0x & 23.23 & 24.16 \\
\midrule
SVD (rank 64) & 3.23x & 25.34 & 26.51 \\
\makecell[l]{Distilled Emb\\(rank 64)} & 3.23x & \textbf{24.88} & \textbf{25.75} \\
\midrule
SVD (rank 32) & 6.47x & 27.06 & 27.91 \\
\makecell[l]{Distilled Emb\\(rank 32)} & 6.47x & \textbf{26.15} & \textbf{27.46} \\
\bottomrule
\end{tabular}
\caption{Language Modeling perplexity for WikiText-103 on validation and test sets. We compressed the embedding matrix from 151M parameters to 34M (3.23x) and 17M (6.47x) parameters. Std is an abbreviation of the word Standard.}
\label{tab:lm}
\end{table}

The Portuguese-English task presents a problem where the embedding matrix constitutes the majority of the parameters of the neural network. The embedding dimension is smaller (256) compared to the other two tasks but embedding compression yields a BLEU score increase in all methods except Structured Embedding. This is due to a regularization effect from the compression. Our model again achieves the highest BLEU score.

On these three experiments we demonstrate that our funneling decomposition method with embedding distillation consistently yields higher BLEU scores compared to existing methods. 

\subsection{Language Modeling}
As a second task we consider language modeling on the WikiText-103 dataset. We compare our method against SVD with two compression rates. The results are presented in Table \ref{tab:lm}. We demonstrate that our distilled embedding method consistently yields lower perplexity (PPL) compared to SVD.

\begin{table}[t]
\centering
\begin{tabular}{lcccc}
\toprule
Model & \makecell{Emb.\\CR} & Init. & \makecell{No\\Distill.} & \makecell{Emb.\\Distill.}  \\ 
\midrule
En-Fr  & 7.87x & Random & 37.04 & 37.21 \\
En-Fr  & 7.87x & Model & 37.54 & \textbf{37.78} \\
\midrule
En-De & 7.89x & Random & 26.07 & 26.35 \\
En-De & 7.89x & Model & 26.7 & \textbf{26.97} \\
\midrule
Pt-En & 3.96x & Random & 42.29 & 42.36 \\
Pt-En & 3.96x & Model & 42.5 & \textbf{42.62}\\
\bottomrule
\end{tabular}
\caption{Comparison of different methods for Funneling (64).}
\label{tab:abl1}
\end{table}

\subsection{Ablation Study}
We present different experiments on machine translation to demonstrate the effect of 1) Model Initialization, 2) Embedding Distillation, 3) Fine-tuning strategies, 4) Compression capability, 5) Alpha Value Sensitivity and 6) Extension and generality of our method.

\paragraph{Initialization} We do an ablation study on all the three language pairs defined in Section \ref{sec:setup}, to conclude, if random initialization is better than model-based initialization. We conclude that model-based initialization, consistently performs better (Table \ref{tab:abl1}).
\paragraph{Embedding Distillation} Table~\ref{tab:abl2} presents different compression rates on the Pt-En task, and embedding distillation performs better across all of them. In Table~\ref{tab:abl1}, we see that across all language pairs when we initialize our model using weights from the funneling decomposition, we improve when using Embedding Distillation during finetuning. We performed embedding distillation with random initialization only on the smaller Pt-En dataset and observed that Embedding Distillation improves BLEU score even with random initialization.
\paragraph{Compression Rate}  We demonstrate in Table~\ref{tab:abl2} that it is possible to compress the embedding up to 15.86x with only a 2\% drop in BLEU score for Pt-En.
\paragraph{Re-training} Fine-tuning is an important component in our method and we demonstrate through our experiments that at convergence most of the techniques are close in performance. Table~\ref{tab:abl3} shows that freezing embedding weights and re-training just the network weights or vice versa leads to a sharp drop in BLEU score, thus, we need to re-train all the weights. The use of a non-linearity and adding embedding distillation also improves BLEU score after finetuning.
\paragraph{Alpha ($\alpha$) Value Sensitivity Analysis} We performed a sensitivity analysis on the $\alpha$ hyper-parameter introduced by our method. Table \ref{tab:abl5} presents our findings. We can see that the method is not very sensitive to the change in $\alpha$ value. We did not tune the alpha for our different experiments but chose the value which gave us good validation results on the WMT En-De translation task. The results of this analysis suggest that we can gain a little performance if we tune alpha for every dataset.
\paragraph{Extension} We experimented with applying two key lessons from our method, namely, using a non-linear function and embedding distillation, to a model initialized with group partitions of the GroupReduce method~\cite{chen2018groupreduce}, we refer to this method as \textit{GroupFunneling}. Table~\ref{tab:abl4} shows that, \textit{GroupFunneling} achieves a higher BLEU score on Pt-En compared to GroupReduce.

\begin{table}[t]
\centering
\begin{tabular}{ccccc}
\toprule
Params &  \makecell{Emb.\\Params} & \makecell{Emb.\\CR} & \makecell{No\\Distill.} & \makecell{Emb.\\Distill.} \\
\midrule
11M & 8M & 1.0x & \textbf{41.43} & - \\
5M & 2M  & 3.96x & 42.50 & \textbf{42.62} \\
4M & 1M  & 7.93x & 42.44 & \textbf{42.60} \\
4M & 516k  & 15.86x & 40.42 & \textbf{40.60}\\
\bottomrule
\end{tabular}
\caption{BLEU scores for different compression rates with bottleneck sizes of 64, 32 and 16 accordingly for IWSLT Pt-En.}
\label{tab:abl2}
\end{table}

\begin{table}[t]
    \centering
    \begin{tabular}{lc}
    \toprule
    Model & BLEU \\
    \midrule
    Proposal  & \textbf{42.60} \\
    \midrule
    - embedding distillation & 42.44 \\
    - non-linearity & 42.34 \\
    \midrule
    Proposal (Freeze non-emb. weights) & 33.34 \\
    Proposal (Freeze emb. weights) & 20.49 \\
    \bottomrule
    \end{tabular}
\caption{BLEU score for IWSLT Pt-En with compression rate 7.93x.}
\label{tab:abl3}
\end{table}

\begin{table}[t]
    \centering
    \begin{tabular}{lc}
    \toprule
    Alpha &  BLEU \\
    \midrule
    0 & 42.50 \\
    0.01 & 42.62 \\
    0.1 & 42.65 \\
    0.3 & 42.66 \\
    0.5 & 42.72 \\
    0.7 & 42.57 \\
    0.9 & 42.03 \\
    \bottomrule
    \end{tabular}
\caption{Alpha value sensitivity analysis on IWSLT Pt-En.}
\label{tab:abl5}
\end{table}

\begin{table}[t]
    \centering
    \begin{tabular}{lc}
    \toprule
    Model &  BLEU \\
    \midrule
    \makecell[l]{GroupFunneling\\(Rand. Initialized + Emb. Distil.) } & \textbf{42.52} \\ \hline
    \makecell[l]{GroupFunneling\\(Rand. Initialized)} & 42.49 \\ \hline
    GroupReduce & 42.13 \\
    \bottomrule
    \end{tabular}
\caption{GroupFunneling (i.e. GroupReduce + Funneling) on IWSLT Pt-En.}
\label{tab:abl4}
\end{table}

\begin{table}[t]
    \centering
    \begin{tabular}{lc}
    \toprule
    Model &  Approx. GFLOPs \\
    \midrule
    SVD & 1.21 \\
    Distilled Embedding & 1.22 \\
    Tensor Train & 2.18 \\
    GroupReduce & 3.41 \\
    \bottomrule
    \end{tabular}
\caption{Approximate GFLOPs on reconstructing the WMT En-De embedding matrix with size $[37000{\times}512]$ and compression rate ~7.89x.}
\label{tab:flops}
\end{table}

\begin{table}[t]
    \centering
    \begin{tabular}{lc}
    \toprule
    Model &  Inference Time (Sec) \\
    \midrule
    Base Model & 27.92 \\
    \midrule
    SVD & 29.63 \\
    Structured Embedding & 31.18 \\
    Distilled Embedding & 29.23 \\
    \bottomrule
    \end{tabular}
\caption{Average inference speed on the IWSLT PT-En model with compression rate ~3.96x.}
\label{tab:time_inf}
\end{table}

\section{Discussion}
\paragraph{Importance of Non-linearity} We postulate that only a subset of word vector dimensions, explains most of the variance, for most word vectors in the embedding matrix. Thus, using ReLU activation might help in regularizing the less important dimensions for a given word vector. 

\paragraph{Importance of Reconstruction Loss} We propose that the embedding reconstruction might suffer from adding the ReLU activation function. The consequence would be loss of information on words not seen during training and loss of generalization performance. Thus, adding a loss for embedding reconstruction helps in grounding the embedding and not lose a lot of information. The amount of regularization is controlled by the hyper-parameter $\alpha$. Our intuition is partly justified by results shown in Table \ref{tab:abl3}, as reconstruction loss performs worse without the ReLU activation function.

\paragraph{Comparison of Inference Speed} We compare the number of floating-point operations used by different models. Table \ref{tab:flops} presents these results. As it is expected, our method is slightly slower than plain SVD method due to the use of the non-linear activation function and the bias additions but notably faster than other more complex methods. Structured embedding does not use any additional floating-point operations, though it requires $groups-1$ additional embedding lookup and concatenate operations. Also, structured embedding requires the reconstruction of the entire embedding matrix at the output projection layer, making it ineffective for model compression.

In addition, we demonstrate on Table \ref{tab:time_inf} the average inference time needed for each method to do a forward pass on the IWSLT Pt-En validation dataset which has a size of 7590 examples. We used a single NVIDIA P100 GPU (12GB) with a batch size of 1024. We averaged the time for 30 runs. We did not perform experiments on GroupReduce and Tensor Train, but according to the Table \ref{tab:flops} we are expecting these methods to be even slower.

\section{Conclusion and future work}
In this paper we proposed Distilled Embedding, a low-rank matrix decomposition with non-linearity in the bottleneck layer for a shared word-embedding and vocabulary projection matrix. We also introduce knowledge distillation of the embedding during fine-tuning using the full embedding matrix as the teacher and the decomposed embedding as the student. We compared our proposed approach with state-of-the-art methods for compressing word-embedding matrix.  We did extensive experiments using three different sizes of datasets and showed that our approach outperforms the state-of-the art methods on the challenging task of machine translation. Our method also generalized well to the task of language modeling. For future work, we will apply our approach to compress feed-forward and multi-head attention layers of the transformer network.

\bibliography{anthology,emnlp2020}
\bibliographystyle{acl_natbib}

\appendix

\section{Appendices}
\label{sec:appendix}

\subsection{Additional Hyper-parameters}
\label{app:hyper}

\paragraph{WMT En-Fr}
Smaller Transformer Network denotes a network with the same configuration as Transformer Base but with hidden size $d_{\text{model}}$ of 416. For GroupReduce, to match the same compression rate we used number of clusters $c$ being equal to 10 and minimum rank $r_{\text{min}}$ to be 22. For SVD, we decided to set the rank to 64. For Tensor Train, we set the embedding shape to be $[25, 32, 40]{\times}[8, 8, 8]$ and the Tensor Train Rank to be 90. For structured embedding we use group size as 32 and number of clusters as 2048, we then use the quantization matrix and learn the clusters from scratch.

\paragraph{WMT En-De}
Smaller Transformer Network denotes a network with the same configuration as Transformer Base but with hidden size $d_{\text{model}}$ of 400. For GroupReduce, to match the same compression rate we used number of clusters $c$ being equal to 10 and minimum rank $r_{\text{min}}$ to be 23. For SVD, we decided to set the rank to 64. For Tensor Train, we set the embedding shape to be $[25, 37, 40]{\times}[8, 8, 8]$ and the Tensor Train Rank to be 90. For structured embedding we use group size as 32 and number of clusters as 2376, we then use the quantization matrix and learn the clusters from scratch.

\begin{table*}[t]
\centering
\begin{tabular}{lcccc}
\toprule
Parameters & Embedding & FFN & Multi-head attention & Linear \\
\midrule
Number & 26M & 25M & 14M & 5M \\
\midrule
Percentage & 37\% & 36\% & 20\% & 7\% \\
\bottomrule
\end{tabular}
\caption{Parameters in the Transformer Base model \cite{NIPS2017_7181} based on a 50k dictionary size and tied input and output embedding.}
\label{tab:param}
\end{table*}

\paragraph{IWSLT Pt-En}
Smaller Transformer Network denotes a network with the same configuration as Transformer Small but with hidden size $d_{\text{model}}$ of 136. For GroupReduce, to match the same compression rate we used number of clusters $c$ being equal to 15 and minimum rank $r_{\text{min}}$ to be 30. For SVD, we decided to set the rank to 64. For Tensor Train, we set the embedding shape to be $[25, 32, 40]{\times}[8, 4, 8]$ and the Tensor Train Rank to be 125. For structured embedding we use group size as 32 and number of clusters as 4048, we then use the quantization matrix and learn the clusters from scratch.

\subsection{Parameter count}

Table~\ref{tab:param} presents the the number of parameters in the different transfomer layers for the transformer base architecture.

\end{document}